\documentclass[preprint,12pt,authoryear]{elsarticle}

\usepackage[utf8]{inputenc}
\usepackage[T1]{fontenc}
\usepackage{amsmath,amssymb,amsfonts}
\usepackage{graphicx}
\usepackage{booktabs}
\usepackage{multirow}
\usepackage{hyperref}
\usepackage{url}
\usepackage{xcolor}
\usepackage{caption}
\usepackage{float}
\usepackage{subfig}
\usepackage{tabularx}
\usepackage{array}
\usepackage{placeins}

\journal{Computers, Environment and Urban Systems}

\begin{document}

\begin{frontmatter}

\title{Cross-View Urban Sensing: Mapping Subjective Streetscape Perception via AlphaEarth Embeddings and Urban Context}

\author[1,2]{Peilin Li}
\ead{liplin25@mail2.sysu.edu.cn}

\author[1,2]{Pengfei Chen\corref{cor1}}
\ead{chenpf9@mail.sysu.edu.cn}

\author[1,2]{Jingyu Wang}
\ead{wangjy595@mail2.sysu.edu.cn}

\author[1,2]{Zhifeng Yang}
\ead{yangzhf23@mail2.sysu.edu.cn}

\author[1,2]{Tiansheng Chen}
\ead{chentsh7@mail2.sysu.edu.cn}

\author[1,2]{Mengjie Gong}
\ead{gongmj@mail2.sysu.edu.cn}

\author[1,2]{Xiao Cheng}
\ead{chengxiao9@mail.sysu.edu.cn}

\cortext[cor1]{Corresponding author}

\affiliation[1]{organization={School of Geospatial Engineering and Science, Sun Yat-sen University, and Southern Marine Science and Engineering Guangdong Laboratory (Zhuhai)}, city={Zhuhai}, postcode={519082}, country={China}}

\affiliation[2]{organization={Key Laboratory of Comprehensive Observation of Polar Environment (Sun Yat-sen University), Ministry of Education}, city={Zhuhai}, postcode={519082}, country={China}}

\begin{abstract}
Residents' perception of the urban streetscape is an important factor in public health, active mobility, and social wellbeing. Street view imagery (SVI) has emerged as a widely used data source for assessing these perceptual qualities, yet its uneven coverage and irregular updating limit large-scale measurement. Here, we present CVLNet, a Cross-View Learning Network that predicts street-level perception from AlphaEarth embeddings and multi-source urban contextual data without requiring SVI at inference. CVLNet applies per-task adaptive gating to jointly model five perceptual dimensions, using labels from the pretrained SVI-Percept model as ground truth.
The proposed method is evaluated across four Southeast Asian cities: Singapore, Kuala Lumpur, Jakarta, and Manila. CVLNet achieves a median road-segment-level Adjusted $R^{2}$ of 0.76 and consistently outperforms the baseline models, with gains ranging from 5.9--11.3\% across the five perceptual dimensions. Ablation experiments show that AlphaEarth features and urban contextual features contribute complementary information. We further produce citywide road-level streetscape perception maps for five subjective perceptual dimensions across all four cities, extending perception estimation from the 13--31\% of the road network directly covered by available SVI to the complete road network of each city. Integrating these maps with WorldPop gridded population data, we quantify exposure inequality across population-density, demographic, and land-use groups using the Deficit Palma Ratio. These results demonstrate that remote sensing can serve as a scalable alternative to SVI for citywide streetscape perception mapping, enabling a more comprehensive assessment of urban environmental inequality.
\end{abstract}

\begin{keyword}
Perception mapping \sep Cross-view learning \sep Street view \sep AlphaEarth \sep Exposure inequality
\end{keyword}

\end{frontmatter}

\section{Introduction}
\label{sec:introduction}

As the primary setting for daily mobility and social interaction, streets play a central role in shaping urban experience \citep{ewing_measuring_2009,sallis_physical_2016}. Subjective streetscape perceptions such as safety and pleasantness influence pedestrian route choice and travel-mode decisions \citep{quercia_aesthetic_2014,lu_effect_2018,zhou_social_2019}, and are associated with residents' wellbeing, physical activity, and social inclusion \citep{salesses_collaborative_2013,dubey_deep_2016,kang_review_2020}. As urban development increasingly emphasises the improvement and renewal of existing urban spaces, large-scale assessment of subjective streetscape perception has become increasingly important for urban planning and management.

Street view imagery (SVI) has emerged as the dominant data source for street-scale environmental assessment because it captures visual information from a pedestrian perspective. It has been widely used to extract objective street indicators, such as the green view index (GVI) and building facade ratio \citep{li_assessing_2015,yan_estimation_2022}. Beyond these physical measurements, SVI also enables the assessment of subjective perceptions by approximating people's visual experience of street environments \citep{dubey_deep_2016,zhang_measuring_2018,kang_review_2020}. Conventional approaches to subjective perception assessment, including questionnaire surveys and expert audits, are limited by sample representativeness and high labour costs, which restrict their application at scale \citep{brownson_measuring_2009}. In response, crowdsourced scoring platforms have substantially expanded the scale and efficiency of perception data collection. The Place Pulse framework \citep{salesses_collaborative_2013} and the StreetScore system \citep{naik_streetscore_2014} were among the first to enable large-scale automated perception scoring by soliciting pairwise comparisons from online participants. Building on these foundations, deep learning models trained on crowdsourced labels have been used to predict perception scores directly from SVI for city-scale mapping \citep{dubey_deep_2016,zhang_social_2019,kruse_places_2021}, and subsequent studies examined their associations with urban vitality and socioeconomic indicators \citep{zhang_measuring_2018,biljecki_street_2021,rzotkiewicz_systematic_2018}. 

Despite these advances, large-scale perception assessment based on SVI is still limited by uneven image coverage and irregular updates. Commercial street view platforms primarily collect imagery along major vehicular roads, leading to sparse or missing coverage in suburban areas and along minor streets \citep{li_assessing_2015,fan_coverage_2025}. Image update frequency also varies considerably, with some urban areas updated every few years, whereas others experience substantially longer intervals between updates \citep{rzotkiewicz_systematic_2018,biljecki_street_2021,wang_assessing_2024}. These limitations are particularly relevant in rapidly changing urban areas, where changes in the built environment can also alter street-level perceptions \citep{liu_physical_2025}.

Considerable effort has been made to overcome the coverage limitation in SVI-based evaluation. One common solution is spatial interpolation. \citet{mooney_street_2017} used Google SVI to perform virtual neighborhood audits and applied spatial interpolation to estimate physical disorder at unobserved locations. Similarly, \citet{liu_establishing_2023} estimated GVI for the complete street network of Shenzhen from sampled street-view observations using ordinary kriging. While the interpolation method is straightforward and easy to implement, its effectiveness depends heavily on the density and spatial distribution of SVI observations, leading to reduced reliability where SVI coverage is sparse. 

Remote sensing data offer a compelling alternative by providing continuous spatial coverage and regular, high-frequency updates across entire city scales \citep{zhu_understanding_2019}. Despite the difference in viewpoint, remote sensing imagery inherently captures physical and spatial patterns that correspond to ground-level environments, as evidenced by recent advances in cross-view synthesis that generate street-view scenes directly from aerial and satellite imagery \citep{regmi_cross-view_2018,xu_satellite_2025,li_bridging_2026}. Building on this premise, \citet{sun_comparison_2026} applied regression models to map street-level GVI using remote sensing vegetation indices and urban form metrics; \citet{pradana_harmonizing_2025} framed the task as a multiclass classification problem, combining remote sensing indices and building footprints with multiple machine-learning classifiers to predict a broader visual taxonomy spanning greenness, enclosure, and openness. Recent advances have also shifted toward deep learning to enhance the predictive performance in complex urban contexts. For example, \citet{ma_first_2025,ma_sky_2025} introduced an end-to-end deep learning framework that predicts GVI directly from satellite image patches, and scaled this predictive pipeline to 19 major Chinese cities, yielding the first cross-city, seasonally resolved GVI dataset. Together, these studies demonstrate that remote sensing can effectively estimate objective streetscape characteristics over large urban areas. However, the capacity of remote sensing to predict subjective streetscape perception remains underexplored.

Here, we propose a \textit{Cross-View Perception Prediction Analysis Framework} for city-scale assessment of subjective streetscape perception using remote sensing observations. It comprises three methodological modules: perception prediction, road-network perception mapping, and population exposure inequality analysis. The contributions are threefold:
\begin{enumerate}
  \item We develop the Cross-View Learning Network (CVLNet), which integrates AlphaEarth (AE) features with multi-source urban contextual (CTX) features through per-task gating to predict five dimensions of street-level perception.
  \item We train CVLNet using ground-truth perception labels derived from sampled SVI and apply it to generate road-level citywide road-level streetscape perception maps in Singapore, Kuala Lumpur, Jakarta, and Manila.
  \item We integrate the perception maps with WorldPop gridded population data to quantify inequalities in perceived urban environments across cities, population-density, demographic, and land-use groups.
\end{enumerate}

\section{Study Area and Data}
\label{sec:data}

\subsection{Study Area}
\label{sec:studyarea}

\begin{figure}[htbp]
  \centering
  \includegraphics[width=0.9\textwidth]{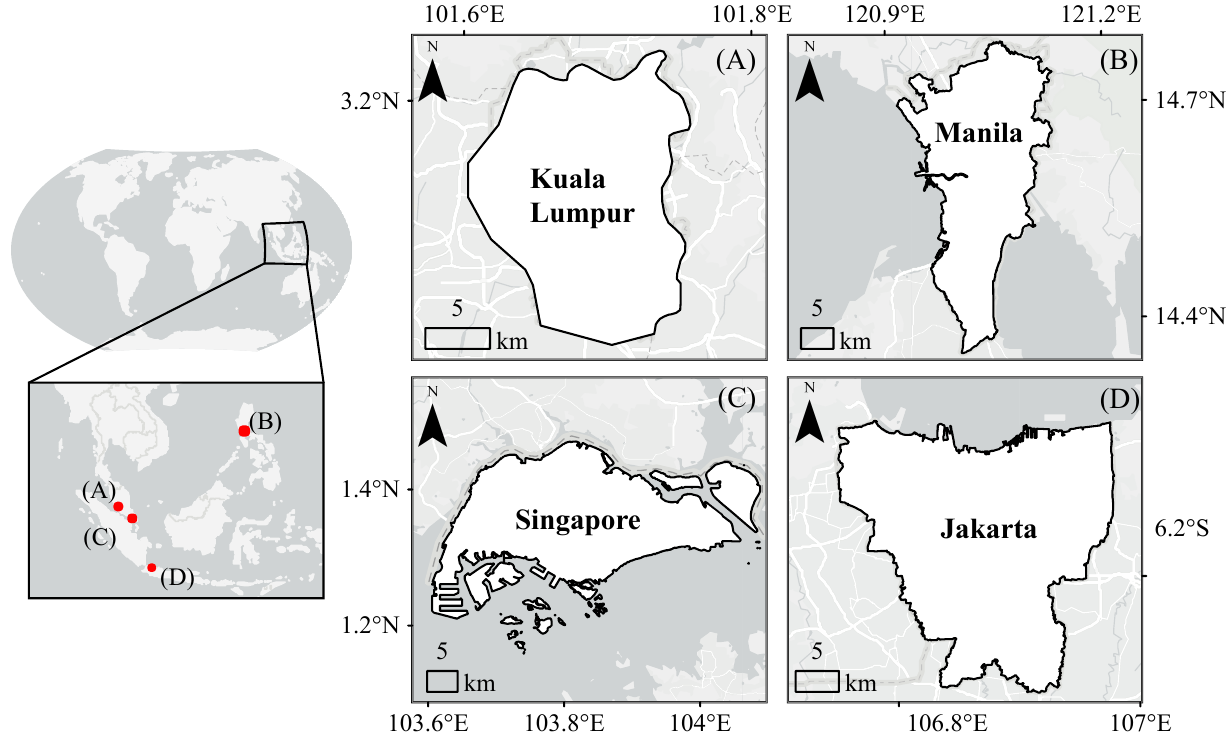}
  \caption{Schematic map of the study area. (A)~Kuala Lumpur, Malaysia; (B)~Manila, Philippines; (C)~Singapore; (D)~Jakarta, Indonesia.}
  \label{fig:studyarea}
\end{figure}

Four Southeast Asian cities, Singapore, Kuala Lumpur, Jakarta, and Manila, are selected as study areas. Their geographic locations, administrative boundaries, and primary road networks are shown in Fig.~\ref{fig:studyarea}. Key city attributes and street view sampling statistics are summarised in Table~\ref{tab:studyarea}. Their tropical setting and comparable streetscape appearance provide a common environmental background for cross-city comparison. Meanwhile, differences in economic development, urban morphology, and built environment characteristics allow our framework to be evaluated across diverse urban contexts.

\begin{table}[htbp]
  \centering
  \caption{Overview of study areas and street view sampling statistics.}
  \label{tab:studyarea}
  \resizebox{\textwidth}{!}{%
  \begin{tabular}{l l r r r r l}
    \toprule
    & \multicolumn{3}{c}{\textbf{City Information}} & \multicolumn{3}{c}{\textbf{SVI Sampling}} \\
    \cmidrule(lr){2-4} \cmidrule(lr){5-7}
    City & Country & Pop.\ (M) & GDP/cap (USD) & Points & Road (km) & Period \\
    \midrule
    Singapore    & Singapore   & 5.8  & 65{,}233 & 84{,}564 & 16{,}474 & 2017--2024 \\
    Kuala Lumpur & Malaysia    & 7.5  & 28{,}364 & 67{,}194 & 16{,}015 & 2017--2024 \\
    Jakarta      & Indonesia   & 10.6 & 12{,}882 & 44{,}170 & 20{,}200 & 2017--2024 \\
    Manila       & Philippines & 13.5 & 8{,}389  & 26{,}547 &  3{,}406 & 2017--2024 \\
    \bottomrule
  \end{tabular}%
  }
  \vspace{2mm}
  \parbox{\textwidth}{\footnotesize Note: Population figures refer to the metropolitan/urban agglomeration level for Kuala Lumpur (Greater KL) and Manila (Metro Manila), and to the city administrative boundary for Singapore and Jakarta (DKI Jakarta).}
\end{table}

\subsection{Study Data}
\label{sec:studydata}

\textbf{Street View Imagery.} Google Street View 360\textdegree{} panoramic imagery is used as the street-level data. The dataset provides georeferenced panoramic imagery along urban road networks, capturing streetscape characteristics such as building facades, vegetation, and road infrastructure.

\textbf{AlphaEarth Features.} AlphaEarth embedding Foundations, developed by Google DeepMind, serve as the core remote sensing data source. This geospatial foundation model is pretrained via self-supervised learning on large-scale multi-temporal Earth observation data, including optical, thermal, and radar satellite imagery together with climate, elevation, and vegetation structure measurements. The model encodes each 10\,m pixel into a 64-dimensional embedding vector that captures multi-spectral surface properties, seasonal dynamics, topographic context, and climatic conditions in a compact latent representation \citep{brown_alphaearth_2025}. The dataset provides annually updated layers covering 2017--2024.

\textbf{Urban Contextual Features.} The urban contextual features used in this study comprise six feature groups: Point of Interest (POI) density features, land-use features, terrain features, socioeconomic features, remote sensing index features, and time-location features. OpenStreetMap, a global open geographic database, provides POI records and land-use polygons \citep{haklay_openstreetmap_2008}. POI records describe facility distribution and activity functions, complementing AlphaEarth features with functional urban semantics \citep{huang_learning_2023}. NASADEM provides elevation data. WorldPop Residential Population Estimates (R2025A v1, 100\,m resolution) and NOAA VIIRS Day/Night Band monthly composites (VNP46A1/VCMSLCFG) provide population and nighttime light data \citep{bondarenko_constrained_2025,roman_black_2018}. Landsat~8 Collection~2 Level~2 provides surface reflectance imagery. The city, geographic coordinates, and image collection year of each sampling point are also retained.

\section{Methods}
\label{sec:methods}

\subsection{Framework Overview}
\label{sec:framework}

\begin{figure}[htbp]
  \centering
  \includegraphics[width=\textwidth]{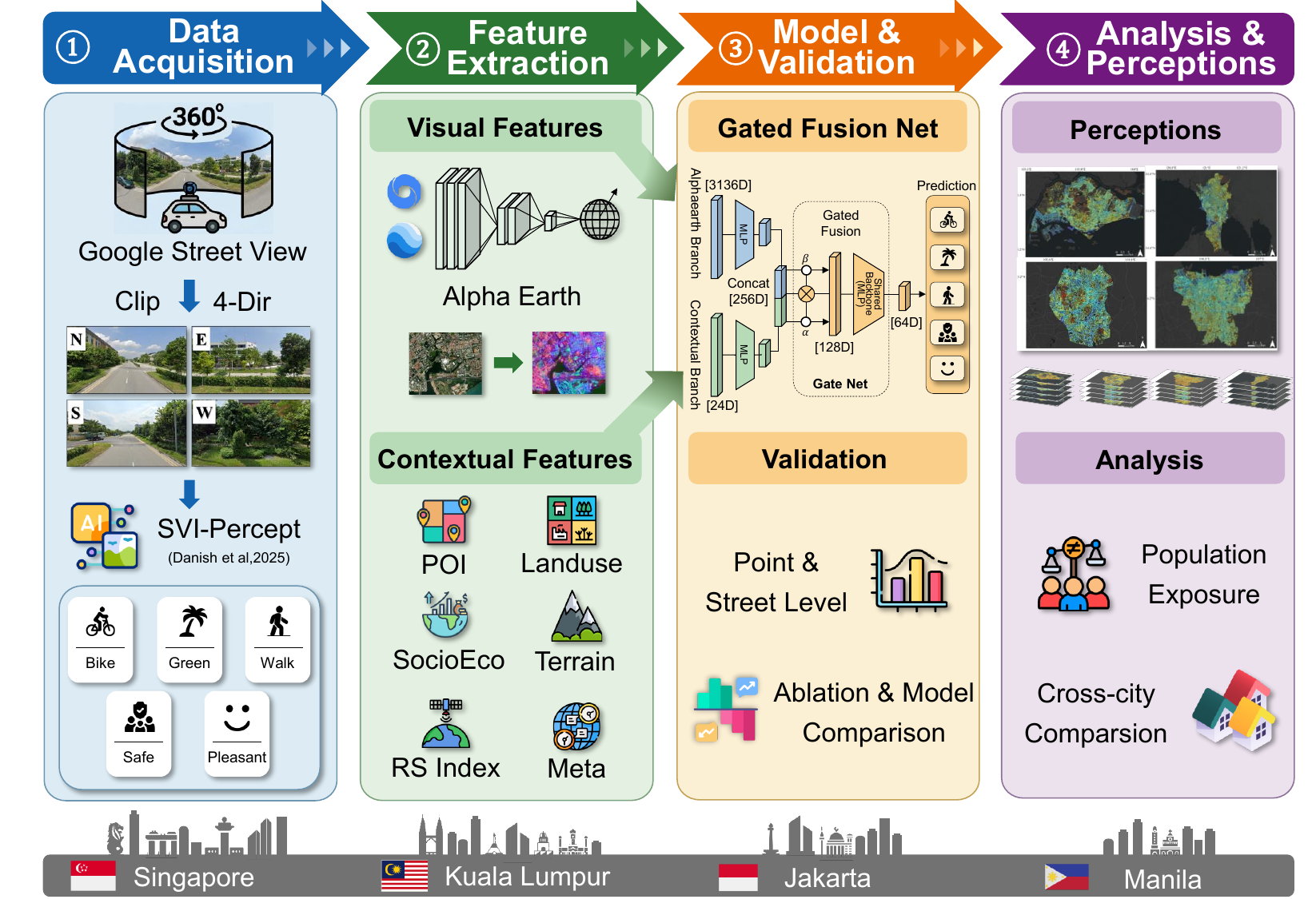}
  \caption{Overview of the Cross-View Perception Prediction Analysis Framework.}
  \label{fig:framework}
\end{figure}

The proposed framework links SVI-derived perception labels, AlphaEarth features, and urban contextual features to predict, map, and analyse subjective streetscape perception at city scale (Fig.~\ref{fig:framework}). The workflow comprises four stages: data acquisition, feature extraction, model training and validation, and perception mapping and inequality analysis.

\subsection{Data Acquisition}
\label{sec:data_acquisition}

Street view sampling is conducted within the administrative boundary of each study city. Sampling points are generated at 100\,m intervals along the road network. A point is retained only where SVI is available.

\begin{figure}[htbp]
  \centering
  \includegraphics[width=\textwidth]{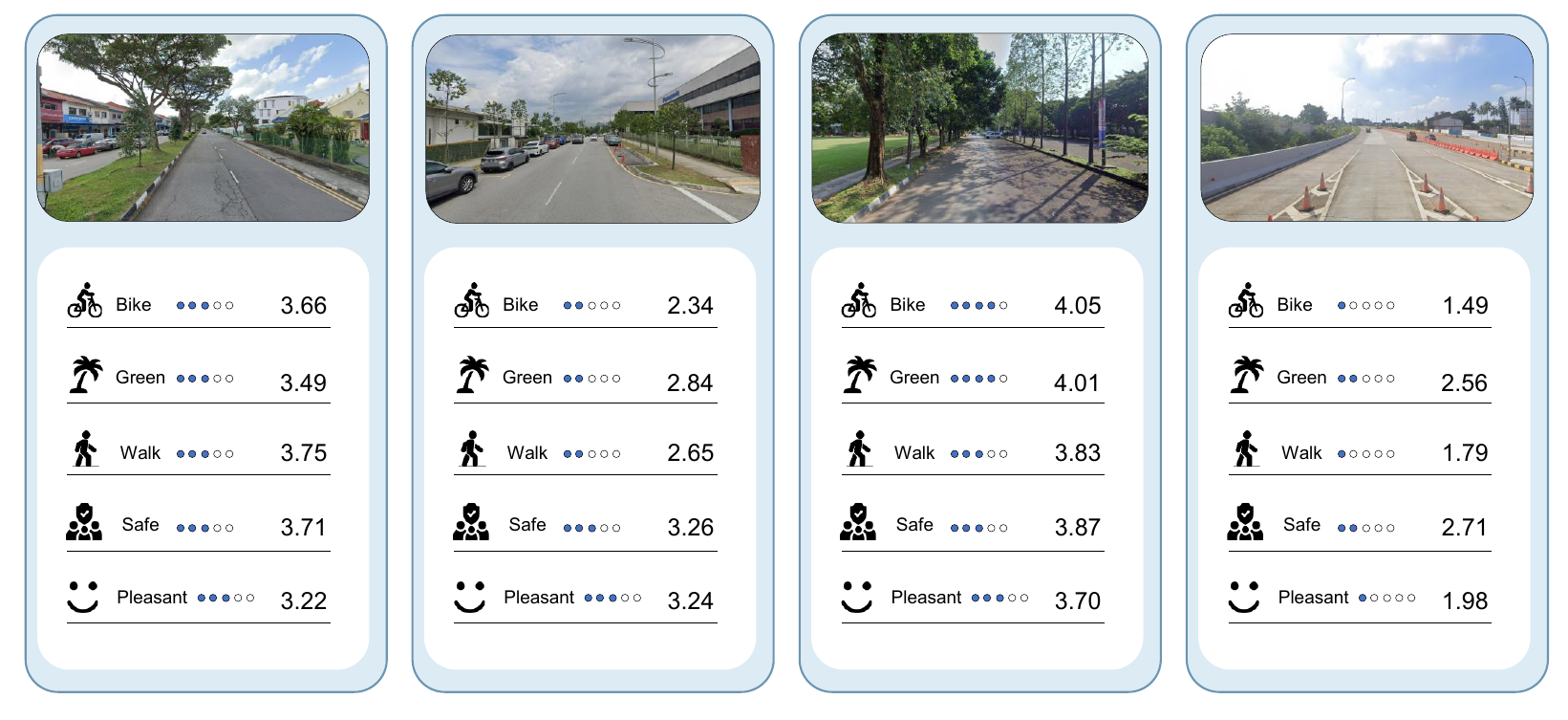}
  \caption{Examples of SVI-Percept scoring for street-view panoramas. The model outputs five perceptual dimensions: bikeability, greenness, pleasantness, safety, and walkability.}
  \label{fig:svi_percept_scoring}
\end{figure}

For model supervision, perception labels are generated from the collected panoramas using the pretrained SVI-Percept model \citep{danish_citizen_2025}. SVI-Percept is a deep learning scoring model developed within a citizen science framework that collected 22{,}637 human ratings from 331 participants for open SVI in Amsterdam. The model predicts five perception scores, namely bikeability, greenness, pleasantness, safety, and walkability, on a 1--5 scale. Each panorama is decomposed into four perspective views using equirectangular projection, with an azimuth interval of 90\textdegree{}, a pitch angle of 0\textdegree{}, and a field of view of 90\textdegree{}. The average score across the four perspective views is used as the panorama-level perception label. Fig.~\ref{fig:svi_percept_scoring} provides four examples of street-view images with their SVI-Percept predicted scores.

\subsection{Feature Extraction}
\label{sec:feature_extraction}

Two feature vectors are constructed for each sampling point: 3{,}136-dimensional AlphaEarth features and 24-dimensional urban contextual features. The AlphaEarth features are formed by extracting a $7\times7$ patch from the annual 64-dimensional embedding layer and flattening the patch. The selection of the patch size is reported in Appendix~\ref{app:patchsize}. The urban contextual features are constructed as follows, with their dimensions and key parameters summarised in Table~\ref{tab:features}.

\textbf{POI Density Features.} POI records are grouped into Food, Shop, Transport, Public, Leisure, and Traffic and converted into continuous density surfaces using Gaussian kernel density estimation with a bandwidth of 2\,km.

\textbf{Land Use Density Features.} Land-use polygons are grouped into Residential, Commercial, Nature, Roads, and Buildings, and the fraction of each category within 2\,km is calculated.

\textbf{Terrain Features.} Elevation in metres and slope in degrees are extracted at each sampling point.

\textbf{Socioeconomic Features.} Population density is matched by city and collection year, and nighttime light intensity is calculated from annual median composites.

\textbf{Remote Sensing Index Features.} The Normalized Difference Vegetation Index (NDVI), Normalized Difference Built-up Index (NDBI), and Modified Normalized Difference Water Index (MNDWI) are calculated from yearly median composites matched to the collection year of each sampling point \citep{tucker_red_1979,zha_use_2003,xu_modification_2006}.

\textbf{Time-Location Features.} These features comprise a city code, sampling point coordinates, and image collection year. The coordinates are the latitude and longitude of each SVI sampling point and are encoded using sine and cosine functions. Collection year is min-max normalised to $[0,\,1]$ over 2017--2024.

Before model training, urban contextual features are standardised using training-set $z$-score statistics, which are retained for testing and inference.

\begin{table}[htbp]
  \centering
  \caption{Summary of multi-source features and key parameters.}
  \label{tab:features}
  \begin{tabular}{llrl}
    \toprule
    Feature Category & Data Source & Dim. & Processing Method \\
    \midrule
    AlphaEarth       & Google AlphaEarth   & 3{,}136 & $7\times7$ patch $\times$ 64D \\
    POI Density      & OpenStreetMap       & 6       & KDE ($\sigma$=2\,km) \\
    Land Use Density & OpenStreetMap       & 5       & 2\,km area ratio \\
    Terrain          & NASADEM             & 2       & Direct sampling \\
    Socioeconomic    & WorldPop / VIIRS    & 2       & Direct sampling \\
    RS Indices       & Landsat 8         & 3       & Yearly median composite \\
    Time-Location   & Sampling Attributes & 6       & Encoding and scaling \\
    \midrule
    \textbf{Total}   & ---                 & \textbf{3{,}160} & --- \\
    \bottomrule
  \end{tabular}
\end{table}

\subsection{Model Training and Validation}
\label{sec:cvlnet}

\subsubsection{CVLNet Architecture}
\label{sec:cvlnet_arch}

\begin{figure}[htbp]
  \centering
  \includegraphics[width=\textwidth]{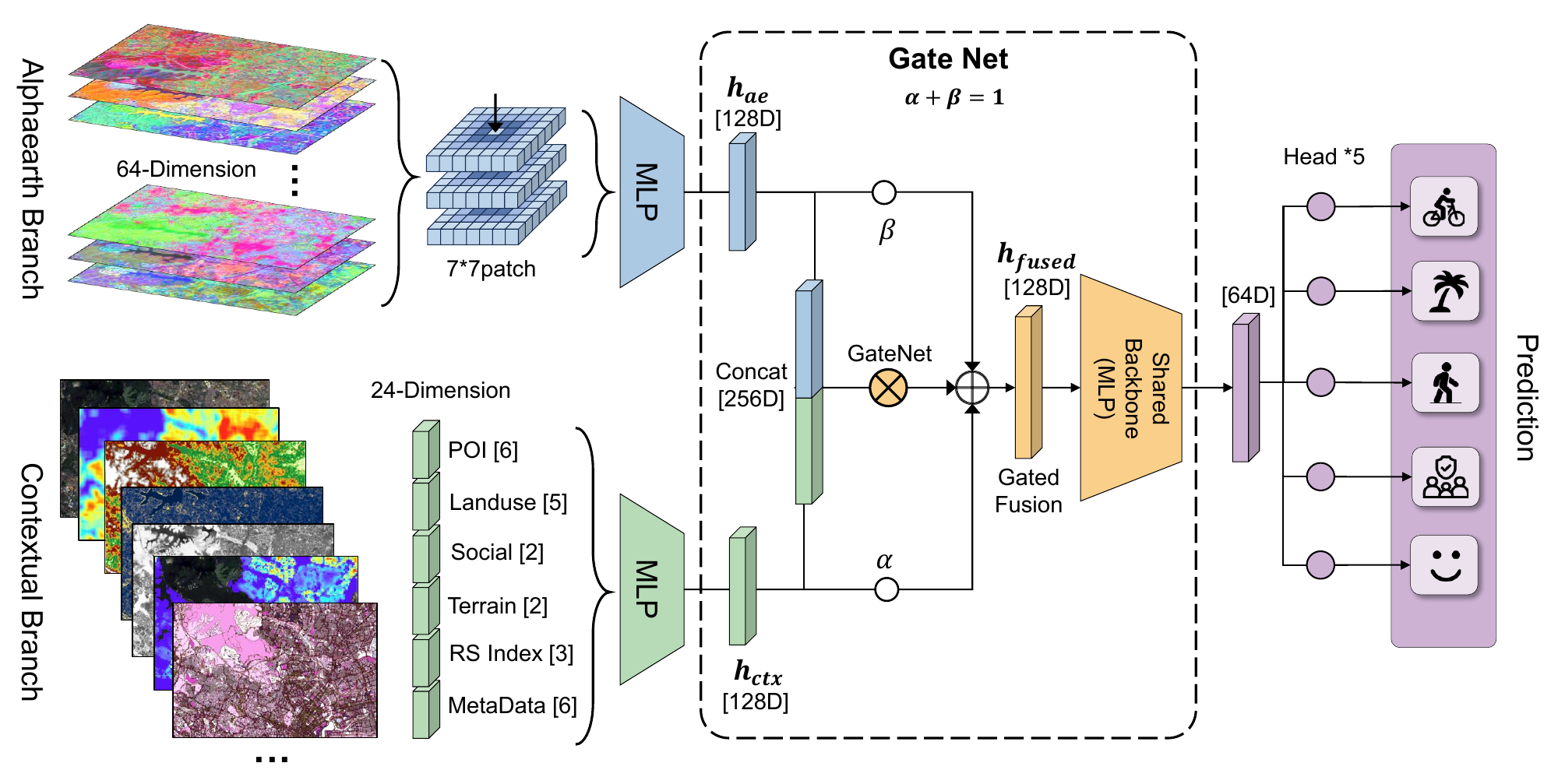}
  \caption{Architecture of the Cross-View Learning Network (CVLNet).}
  \label{fig:CVLNet}
\end{figure}

CVLNet is designed to predict five street-level perceptual dimensions from two feature sources: AE and CTX features. Its architecture has three main components: a dual-branch encoder, a feature fusion module, and a multi-task prediction module (Fig.~\ref{fig:CVLNet}). The dual-branch design allows AE and CTX features to be represented separately before they are combined.

First, separate multilayer perceptrons map the AE and CTX features into a common 128-dimensional embedding space, yielding $\mathbf{h}_{\mathrm{ae}}$ and $\mathbf{h}_{\mathrm{ctx}}$, respectively. This step reduces the dimensional mismatch between the AE branch and the CTX branch. The fusion module then combines the two branch representations for each perceptual target. For the $k$-th target ($k = 1, \ldots, 5$), the gate estimates a fusion weight from the concatenated branch representations, as shown in Eq.~\ref{eq:gate}:

\begin{equation}
  \alpha_k = \sigma\!\left(\mathrm{Linear}_{64 \to 1}\!\left(\mathrm{GELU}\!\left(\mathrm{Linear}_{256 \to 64}\!\left([\mathbf{h}_{\mathrm{ae}};\, \mathbf{h}_{\mathrm{ctx}}]\right)\right)\right)\right)
  \label{eq:gate}
\end{equation}
\begin{equation}
  \mathbf{h}_{\mathrm{fused}}^{(k)} = \alpha_k \cdot \mathbf{h}_{\mathrm{ae}} + (1 - \alpha_k) \cdot \mathbf{h}_{\mathrm{ctx}}
  \label{eq:fusion}
\end{equation}
where $\sigma(\cdot)$ is the sigmoid function and $\alpha_k \in [0,\,1]$ controls the relative weight assigned to the AE branch. The fused representation is then computed using Eq.~\ref{eq:fusion}. A larger $\alpha_k$ indicates that the trained model relies more on AE features for that perceptual target, whereas a smaller value indicates greater reliance on CTX features. This value is used as a coarse diagnostic signal for comparing feature-source reliance across perceptual dimensions. The fused representation $\mathbf{h}_{\mathrm{fused}}^{(k)}$ is passed to a shared prediction backbone, and separate output heads produce the five perception scores.

\subsubsection{Model Training Configuration}
\label{sec:config}

CVLNet is trained as a multi-task regression model. In each forward pass, it predicts five perception scores: bikeability, greenness, pleasantness, safety, and walkability. Let $y$ denote the observed perception score and $\hat{y}$ denote the predicted score. We use Huber loss with $\delta=1.0$ (Eq.~\ref{eq:huber}) instead of MSE because it gives a linear penalty to large errors \citep{huber_robust_1964}:
\begin{equation}
  \mathcal{L}_{\text{Huber}}(y, \hat{y}) =
  \begin{cases}
    \frac{1}{2}(y-\hat{y})^2, & |y-\hat{y}| \leq \delta \\[4pt]
    \delta\cdot|y-\hat{y}| - \frac{1}{2}\delta^2, & |y-\hat{y}| > \delta
  \end{cases}
  \label{eq:huber}
\end{equation}
For the $k$-th perceptual dimension, $y_k$ and $\hat{y}_k$ denote the observed and predicted scores, respectively. The total loss sums the Huber losses over the five perceptual dimensions with equal weights:
\begin{equation}
  \mathcal{L} = \sum_{k=1}^{5} \mathbf{1}_{[\text{valid}_k]} \cdot \mathcal{L}_{\text{Huber}}(y_k, \hat{y}_k)
  \label{eq:totalloss}
\end{equation}
where $\mathbf{1}_{[\text{valid}_k]}$ is an indicator that equals 1 when the $k$-th target is available for a sample and 0 otherwise. Equal weighting is adopted as a baseline design; task-specific weighting strategies may further improve performance on harder dimensions. The model is optimised with Adam at an initial learning rate of $1 \times 10^{-3}$ and a batch size of 512 for up to 80 epochs. A cosine annealing schedule and early stopping are used to stabilise convergence. Dropout is applied to both branches.

\subsubsection{Spatially Stratified Data Splitting}
\label{sec:splitting}

The data in each city are split using a regular 2\,km $\times$ 2\,km grid, which is intended to reduce spatial data leakage between the training and test sets \citep{roberts_cross-validation_2017}. Within each city, 80\% of the grid cells are randomly selected for training and the remaining 20\% for testing. The training data from the four cities are combined to train a global model. The test data are retained separately for each city.

\subsubsection{Evaluation Metrics}
\label{sec:metrics}

Model performance is evaluated using adjusted coefficient of determination (Adj.\ $R^{2}$) and normalised root mean squared error (nRMSE). Adj.\ $R^{2}$ measures the proportion of variance explained by the model while accounting for the number of predictors. The standard coefficient of determination is first computed as:
\begin{equation}
  R^2 = 1 - \frac{\sum_{i=1}^{N}(y_i - \hat{y}_i)^2}{\sum_{i=1}^{N}(y_i - \bar{y})^2}
  \label{eq:r2}
\end{equation}
where $y_i$ is the observed perception score, $\hat{y}_i$ is the predicted score, and $\bar{y}$ is the mean observed score. The adjusted $R^{2}$ is then computed as:
\begin{equation}
  R^2_{\text{adj}} = 1 - (1 - R^2) \cdot \frac{N - 1}{N - p - 1}
  \label{eq:adjr2}
\end{equation}
where $N$ denotes the sample size and $p$ denotes the number of predictors. 

Prediction error is evaluated using nRMSE, which divides RMSE by the theoretical range of the perception scores:
\begin{equation}
  \text{nRMSE} = \frac{\sqrt{\frac{1}{N}\sum_{i=1}^{N}(y_i - \hat{y}_i)^2}}{y_{\max} - y_{\min}}
  \label{eq:nrmse}
\end{equation}
where $y_{\max}=5$ and $y_{\min}=1$ are the theoretical maximum and minimum perception scores. The RMSE is divided by $y_{\max} - y_{\min}=4.0$ to obtain a normalised error.

Model performance is reported at both point and road-segment levels. Road-segment-level evaluation is conducted for road segments with at least three test points. For each eligible road segment, the observed and predicted point-level perception scores are averaged separately to obtain the observed and predicted road-segment-level perception scores. Adjusted $R^{2}$ and nRMSE are then computed using the eligible road segments as the evaluation units.

\subsubsection{Baseline Model Comparison and Ablation Analysis}
\label{sec:baseline}

For the model comparison, five baseline models from two categories are evaluated: the RandomForest and XGBoost baselines as machine-learning models, and the D-MLP, ResNet, and Transformer baselines as neural-network models. For both machine-learning baselines, the flattened AE features are concatenated with the CTX features and used as a single input. The RandomForest baseline uses 100 estimators \citep{breiman_random_2001}. The XGBoost baseline uses 200 estimators and a learning rate of 0.05 \citep{chen_xgboost_2016}.

For the neural-network comparison, the AE and CTX features follow the same branch-based processing and multi-task output setting as CVLNet, while the network applied after feature fusion is changed. The D-MLP baseline uses two fully connected layers with 256 and 128 units. The ResNet baseline maps the fused features to 128 units, applies two residual blocks, and reduces the output to 64 units before prediction \citep{he_deep_2016}. The Transformer baseline divides the fused features into eight 32-dimensional tokens and uses two encoder layers with four attention heads and a feed-forward dimension of 128 \citep{vaswani_attention_2017}. All three baselines use separate output heads for the five perceptual dimensions and follow the training configuration used for CVLNet.

Two single-branch ablation variants are evaluated against the full CVLNet model: an AE-only model using only AE features and a CTX-only model using only CTX features. In the single-branch variants, the gating module is removed while the remaining architecture and training settings are kept unchanged. This comparison is used to assess how each input branch affects prediction performance.

\subsection{Perception Mapping and Inequality Analysis}
\label{sec:inequality}

\subsubsection{Perception Mapping}
\label{sec:perception_mapping}

The trained CVLNet model is applied to generate citywide road-level streetscape perception maps for the five perceptual dimensions. For each prediction location, the model extracts $7\times7$ AE features from the annual 10\,m AlphaEarth embedding layers and combines them with the corresponding CTX features. The point-level predictions are assembled into road-network-based perception maps and exported as five maps for each city, one for each perceptual dimension.

\subsubsection{Population Exposure Inequality Analysis}

Following established approaches for measuring distributional inequalities in urban amenities and burdens \citep{logan_measuring_2021}, population exposure is calculated by linking each population point to the predicted perception score at the same location. For dimension $k$ and population point $i$, $S_{k,i}$ denotes the predicted perception score and $P_i$ denotes the population count from WorldPop. Mean exposure is calculated as the population-weighted average of predicted perception scores within each analysis group. The analysis covers all population points in each city, demographic groups, and land-use groups, including Residential, Commercial, and Nature zones.

The Deficit Palma Ratio is used as a descriptive inequality measure. For each dimension, the perception deficit of point $i$ is defined as $D_{k,i} = S_{\max} - S_{k,i}$, where $S_{\max} = 5.0$. Population points are ordered by $D_{k,i}$. Population counts are then used to identify the 10\% of the population in points with the highest deficits and the 40\% in points with the lowest deficits. The ratio is defined as:
\begin{equation}
  \mathrm{Deficit\ Palma}_k = \frac{\mathrm{Share}_{D,\,\mathrm{top}\,10\%}}{\mathrm{Share}_{D,\,\mathrm{bottom}\,40\%}}
  \label{eq:palma}
\end{equation}
In Eq.~\ref{eq:palma}, the numerator is the share of total population-weighted deficit for the highest-deficit 10\% group. The denominator is the corresponding share for the lowest-deficit 40\% group. Under a proportional distribution, these two shares are 10\% and 40\%, giving a reference value of 0.25. Values above 0.25 indicate a higher concentration of deficits in the highest-deficit group.

Exposure differences across population density levels are examined using population-density quintiles. Within each city, population points are ordered by population count and divided into five equal-sized groups, from the sparsest 20\% to the densest 20\%. A population-weighted mean perception score is calculated for each group. The difference between the densest and sparsest groups is defined as:
\begin{equation}
  \Delta_{k} = \bar{S}_k^{(G_5)} - \bar{S}_k^{(G_1)}
  \label{eq:delta}
\end{equation}
In Eq.~\ref{eq:delta}, $\bar{S}_k^{(G_5)}$ and $\bar{S}_k^{(G_1)}$ denote the population-weighted mean scores of the densest and sparsest groups. A negative $\Delta_k$ indicates a lower population-weighted mean score in the densest group than in the sparsest group.

\section{Results}
\label{sec:results}

\subsection{Model Performance}
\label{sec:performance}

\subsubsection{Model Prediction Accuracy}
\label{sec:ModelAccuracy}

\begin{figure}[htbp]
  \centering
  \includegraphics[width=0.9\textwidth]{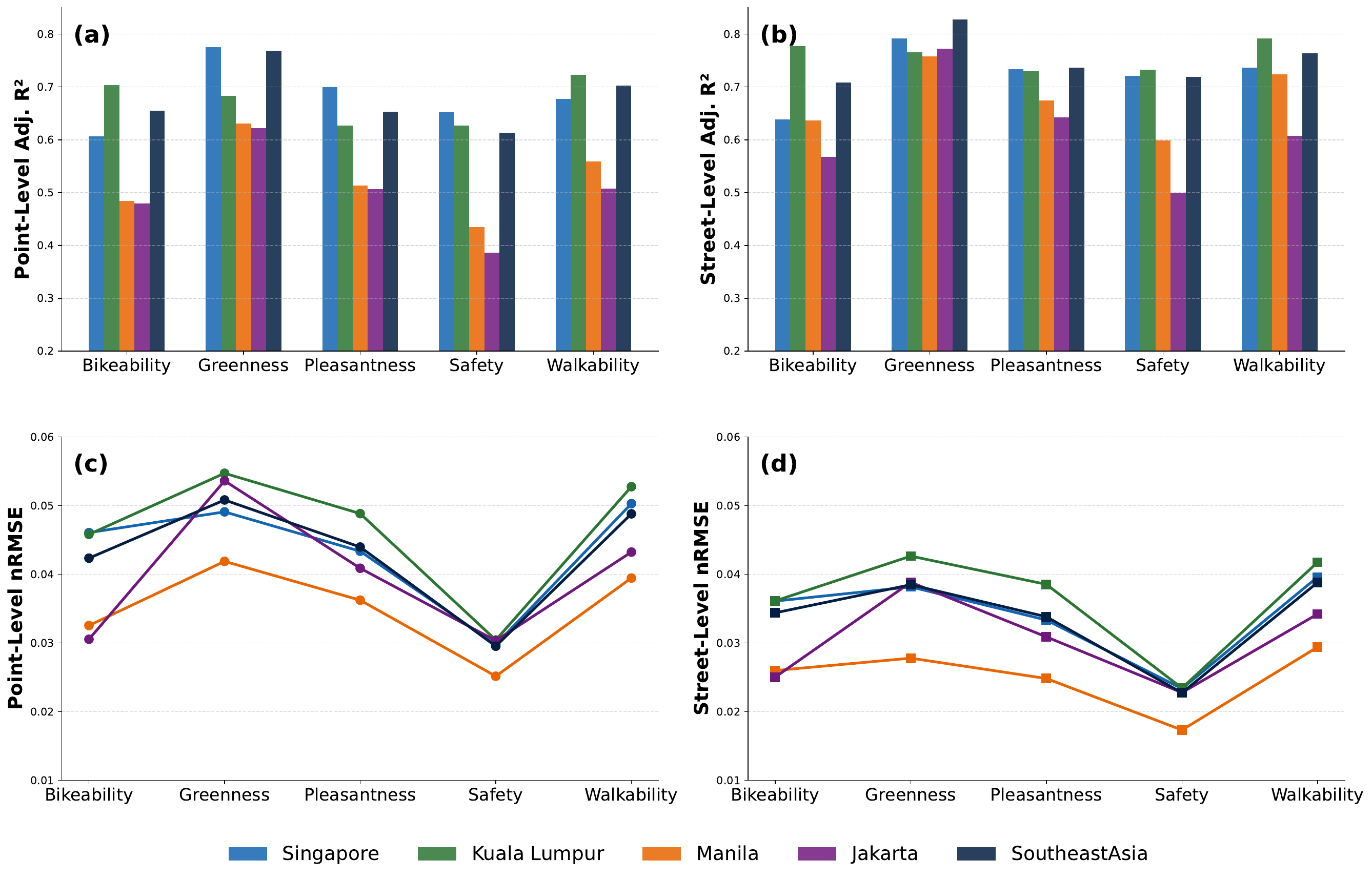}
  \caption{Performance evaluation of CVLNet across multiple Southeast Asian cities. (a)--(b): Adjusted $R^{2}$ values measured at the point and road-segment levels; (c)--(d): normalised root mean squared error (nRMSE).}
  \label{fig:performance}
\end{figure}

Fig.~\ref{fig:performance} presents the point-level and road-segment-level prediction accuracy of CVLNet across the four Southeast Asian cities.

At the point level (Fig.~\ref{fig:performance}a, c), CVLNet shows positive Adj.\ $R^{2}$ values across all five perceptual dimensions. Greenness has the highest Adj.\ $R^{2}$ of 0.768, followed by walkability, while safety has the lowest Adj.\ $R^{2}$ of 0.613. Pleasantness and bikeability fall between these two ends. The nRMSE values are 0.07 or lower across all dimensions, and safety has the lowest nRMSE of 0.04. This combination indicates that safety has relatively small absolute prediction error but lower explained variance than the other dimensions. Across cities, Singapore and Kuala Lumpur generally show higher prediction accuracy than Jakarta and Manila. Singapore has higher values for greenness, pleasantness, and safety, while Kuala Lumpur has the highest values for walkability and bikeability. Jakarta and Manila show lower accuracy in several dimensions, with larger differences appearing for safety and bikeability.

At the road-segment level (Fig.~\ref{fig:performance}b, d), predictions are aggregated using the procedure described in Section~\ref{sec:metrics}, yielding 8{,}273 eligible road segments. In the four-city aggregate results, road-segment aggregation produces higher Adj.\ $R^{2}$ than point-level evaluation for all five dimensions, with nRMSE below 0.04. The median road-segment-level Adj.\ $R^{2}$ is 0.76. Road-segment aggregation substantially improves model accuracy across all dimensions. Averaging multiple point-level perception scores within each eligible road segment reduces the influence of local point-level variation. Overall, these results indicate reliable prediction performance at the road-segment level.

The learned gate weights are reported in Table~\ref{tab:gate_weights}. The gate weight controls the relative coefficients assigned to the AE and CTX branches in the fusion module. Overall mean gate weights range from 0.586 for safety to 0.636 for greenness, indicating that AE receives a moderately larger average coefficient while CTX remains part of the fused representation. The weights vary across cities and perceptual dimensions. Jakarta records the highest mean AE weight in all five dimensions, with the largest cross-city differences observed for safety and pleasantness. The within-city standard deviations further indicate that the learned weights vary among samples. Together, these results support the adaptability of CVLNet and the effectiveness of $\alpha_k$ in adjusting the contributions of AE and CTX features.

\begin{table}[htbp]
  \centering
  \caption{Gate weights of the AE branch by city and perceptual dimension. Values are reported as mean $\pm$ standard deviation.}
  \label{tab:gate_weights}
  \resizebox{\textwidth}{!}{%
  \begin{tabular}{lccccc}
    \toprule
    Perceptual dimension & Overall & Singapore & Kuala Lumpur & Manila & Jakarta \\
    \midrule
    Bikeability  & $0.598 \pm 0.093$ & $0.584 \pm 0.108$ & $0.605 \pm 0.087$ & $0.600 \pm 0.074$ & $0.621 \pm 0.064$ \\
    Greenness    & $0.636 \pm 0.097$ & $0.631 \pm 0.099$ & $0.602 \pm 0.078$ & $0.653 \pm 0.095$ & $0.702 \pm 0.093$ \\
    Pleasantness & $0.607 \pm 0.089$ & $0.604 \pm 0.088$ & $0.594 \pm 0.070$ & $0.561 \pm 0.102$ & $0.674 \pm 0.075$ \\
    Safety       & $0.586 \pm 0.123$ & $0.543 \pm 0.134$ & $0.575 \pm 0.101$ & $0.635 \pm 0.090$ & $0.680 \pm 0.089$ \\
    Walkability  & $0.589 \pm 0.103$ & $0.565 \pm 0.118$ & $0.589 \pm 0.091$ & $0.611 \pm 0.069$ & $0.632 \pm 0.089$ \\
    \bottomrule
  \end{tabular}%
  }
\end{table}

\subsubsection{Baseline Model Comparison and Ablation Analysis}
\label{sec:ablation_results}

\begin{figure}[htbp]
  \centering
  \includegraphics[width=0.9\textwidth]{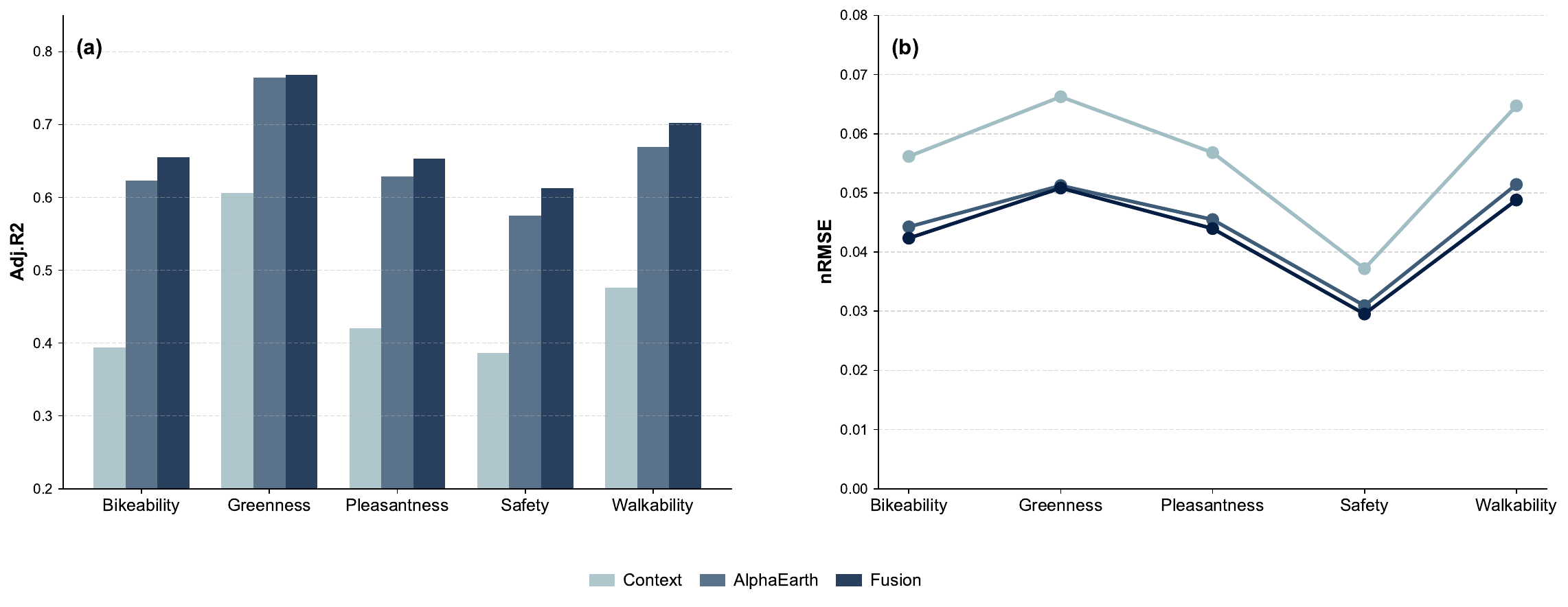}
  \caption{Ablation study of CVLNet and performance comparison against baseline models. (a)(c)~Adjusted $R^{2}$ for ablation variants and comparative models; (b)(d)~corresponding nRMSE traces.}
  \label{fig:ablation}
\end{figure}

The ablation analysis (Fig.~\ref{fig:ablation}) compares the full CVLNet model with AE-only and CTX-only variants. The AE-only model retains substantial predictive accuracy, with Adj.\ $R^{2}$ values ranging from 0.575 to 0.764 across the five perceptual dimensions. This result shows that AE features alone capture a large share of the variation in the SVI-derived perception labels. When only CTX features are used, the model no longer includes the direct remote sensing representation and achieves lower prediction accuracy than the AE-only model in most dimensions, indicating that AE features are an important source of predictive information in CVLNet. The full CVLNet model shows an overall improvement over both single-branch variants across the five dimensions. These results show that combining AE and CTX features through per-task gating improves prediction beyond either feature source alone.

Table~\ref{tab:comparison} further compares CVLNet with five baseline models at the point level. CVLNet achieves the highest Adj.\ $R^{2}$ and the lowest nRMSE in all five perceptual dimensions. Compared with the best non-CVLNet baseline, the largest Adj.\ $R^{2}$ gains are 0.036 for walkability, 0.034 for bikeability, 0.034 for safety, and 0.033 for pleasantness. The gain for greenness is relatively smaller, at 0.009, which is consistent with the relatively high baseline accuracy for this dimension. Overall, CVLNet provides the most consistent performance across the five perceptual dimensions, with higher explained variance and lower prediction error than the baseline models.

\begin{table}[htbp]
  \centering
  \caption{Point-level prediction performance of CVLNet and baseline models.}
  \label{tab:comparison}
  \resizebox{\textwidth}{!}{%
  \begin{tabular}{lccccc}
    \toprule
    \multicolumn{6}{c}{\textbf{Adjusted $R^{2}$}} \\
    \midrule
    Model & Bikeability & Greenness & Pleasantness & Safety & Walkability \\
    \midrule
    RandomForest & 0.5887 & 0.7361 & 0.5936 & 0.5546 & 0.6444 \\
    XGBoost      & 0.6079 & 0.7579 & 0.6160 & 0.5646 & 0.6588 \\
    D-MLP        & 0.6181 & 0.7582 & 0.6195 & 0.5682 & 0.6610 \\
    ResNet       & 0.6108 & 0.7540 & 0.6204 & 0.5788 & 0.6645 \\
    Transformer  & 0.6210 & 0.7592 & 0.6149 & 0.5652 & 0.6658 \\
    \textbf{CVLNet (ours)} & \textbf{0.6553} & \textbf{0.7682} & \textbf{0.6529} & \textbf{0.6129} & \textbf{0.7022} \\
    \midrule
    \multicolumn{6}{c}{\textbf{nRMSE}} \\
    \midrule
    Model & Bikeability & Greenness & Pleasantness & Safety & Walkability \\
    \midrule
    RandomForest & 0.0551 & 0.0644 & 0.0570 & 0.0379 & 0.0641 \\
    XGBoost      & 0.0545 & 0.0638 & 0.0562 & 0.0380 & 0.0630 \\
    D-MLP        & 0.0546 & 0.0640 & 0.0563 & 0.0381 & 0.0631 \\
    ResNet     & 0.0548 & 0.0641 & 0.0567 & 0.0385 & 0.0628 \\
    Transformer  & 0.0555 & 0.0641 & 0.0571 & 0.0387 & 0.0636 \\
    \textbf{CVLNet (ours)} & \textbf{0.0529} & \textbf{0.0635} & \textbf{0.0550} & \textbf{0.0369} & \textbf{0.0610} \\
    \bottomrule
  \end{tabular}%
  }
\end{table}

\subsection{Citywide Perception Mapping}
\label{sec:mapping}

\begin{figure}[htbp]
  \centering
  \includegraphics[width=\textwidth]{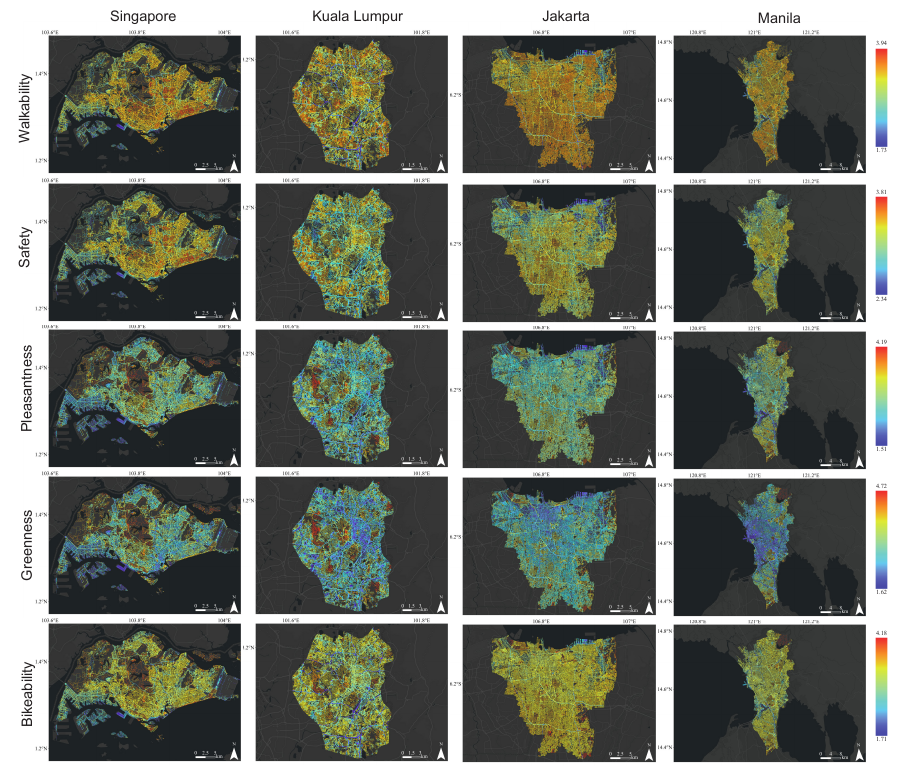}
  \caption{Citywide mapping results of five perceptual dimensions, shown from top to bottom as walkability, safety, pleasantness, greenness, and bikeability, for Singapore, Kuala Lumpur, Jakarta, and Manila.}
  \label{fig:mapping}
\end{figure}

As shown in Appendix~\ref{app:svi_coverage}, the retrieved Google Street View samples cover only part of the road networks in the four cities, with notable cross-city differences in coverage. Consequently, the available SVI samples alone do not provide perception estimates for the complete citywide road networks. Applying the model described in Section~\ref{sec:perception_mapping}, we obtain continuous perception maps across all four cities for the five perceptual dimensions (Fig.~\ref{fig:mapping}).

Figure~\ref{fig:mapping} shows citywide road-level streetscape perception maps for the four cities. The same city can show different spatial distributions among the five perceptual dimensions. Each row uses a unified colourbar for the four cities. Walkability and bikeability show more continuous medium-to-high scores in Jakarta, while Manila is mainly characterised by medium scores with fewer high-score clusters. Singapore and Kuala Lumpur show stronger local heterogeneity, with high- and low-score road segments interlaced across short distances. Safety is less visually polarised than greenness and pleasantness. Singapore contains more visible high-score safety clusters, whereas Kuala Lumpur, Jakarta, and Manila are dominated by medium scores with scattered low-score segments. Greenness and pleasantness show broader low-value areas in Jakarta and Manila, while Singapore and Kuala Lumpur contain sharper local contrasts.

\begin{figure}[htbp]
  \centering
  \includegraphics[width=\textwidth]{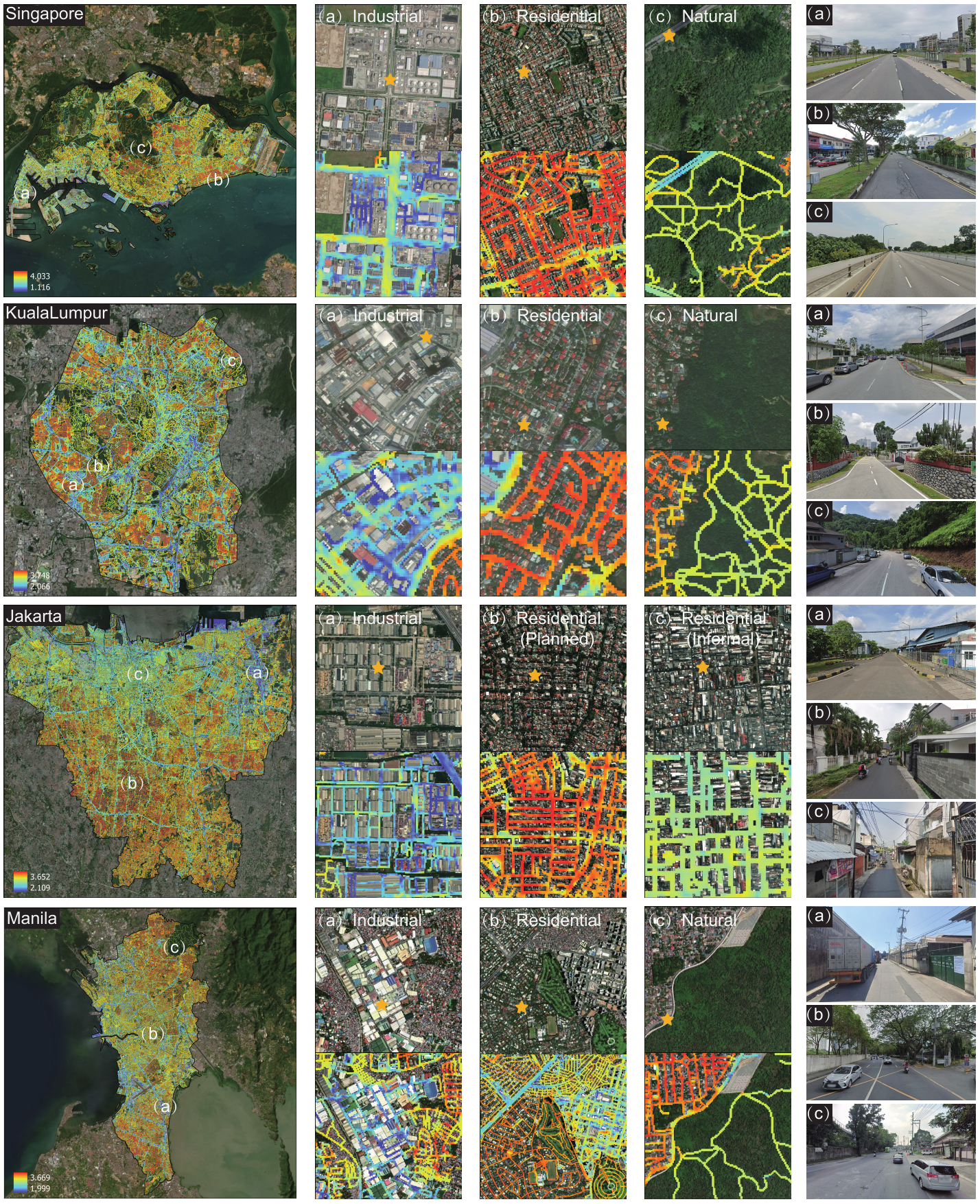}
  \caption{Citywide and local-scale maps of predicted safety perception in Singapore, Kuala Lumpur, Jakarta, and Manila. The stars mark the locations shown in the corresponding street-view images in the far-right column.}
  \label{fig:detail}
\end{figure}

Safety is selected for the detailed maps because it is a representative dimension in street-view perception research and is closely related to everyday mobility and public-space use \citep{naik_streetscore_2014,dubey_deep_2016}. It also represents a composite perceptual dimension that covers multiple aspects of subjective streetscape perception. Figure~\ref{fig:detail} compares detailed mapping results for several urban settings. In the selected cases, industrial areas generally have lower safety scores than residential areas, while the natural areas in Singapore, Kuala Lumpur, and Manila are mainly characterised by medium scores. High-score road segments are more continuous in the residential areas of Singapore (first row, b) and Kuala Lumpur (second row, b), whereas the residential area in Manila (fourth row, b) shows a more mixed distribution. In Jakarta (third row, b--c), the planned residential area has a more regular road and block arrangement and contains more high-score road segments, whereas the denser and less regular informal residential area is mainly characterised by medium scores. Although both are residential areas, their predicted safety score distributions differ.

\subsection{Population Exposure Inequality}
\label{sec:inequality_results}

\begin{figure}[htbp]
  \centering
  \includegraphics[width=0.9\textwidth]{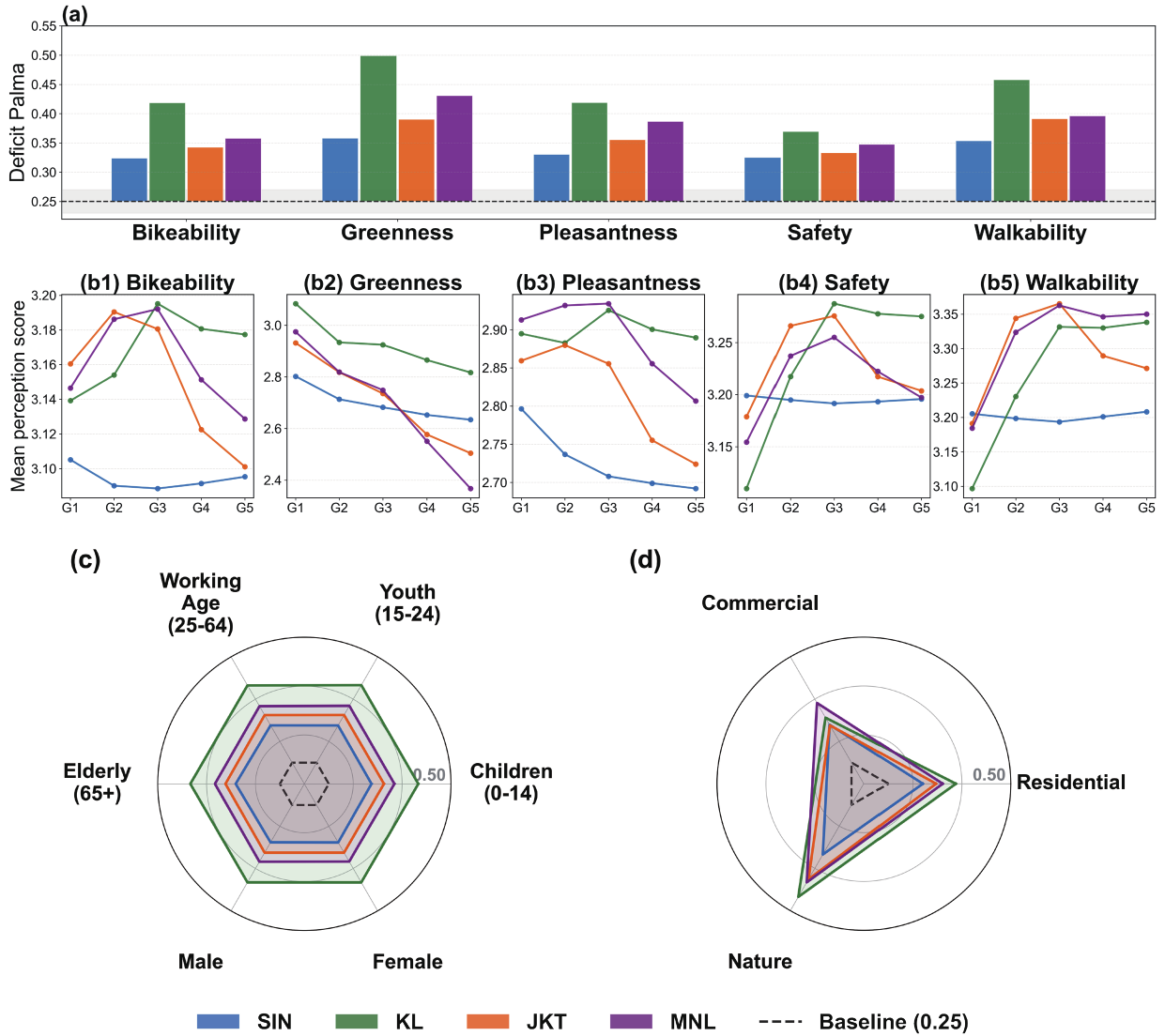}
  \caption{Cross-city exposure inequality analysis. (a)~Deficit Palma Ratio values across cities and perceptual dimensions, with the dashed line marking the 0.25 baseline; (b1)--(b5)~population-weighted mean perception scores across five population-density quintiles ($G_1$ = sparsest 20\%, $G_5$ = densest 20\%) for each perceptual dimension; (c)~Deficit Palma Ratio values by demographic group; (d)~Deficit Palma Ratio values by land-use group.}
  \label{fig:inequality}
\end{figure}

Fig.~\ref{fig:inequality} first reports overall population exposure inequality using the Deficit Palma Ratio and then examines population-density, demographic, and land-use groups.

The overall Deficit Palma Ratio values (Fig.~\ref{fig:inequality}a) show that exposure deficits are unevenly distributed in every city and dimension, with all bars above the equity baseline. Kuala Lumpur consistently has the highest deficit concentration, while Singapore is generally the lowest. Jakarta and Manila occupy intermediate positions, although their relative levels vary by perceptual dimension. Across dimensions, greenness and walkability show stronger deficit concentration than safety.

Across the population-density quintiles (Fig.~\ref{fig:inequality}b1--b5), greenness shows the clearest common pattern. Mean greenness decreases from sparsely populated to densely populated groups in all four cities. Pleasantness shows a similar but weaker decline in most cities. Walkability and safety do not follow the same density gradient. Walkability remains relatively high in the denser groups for Kuala Lumpur, Jakarta, and Manila, while Singapore changes little across density groups. Bikeability shows a more mixed pattern, with mid-density groups often scoring higher than the sparsest and densest groups.

The demographic comparison (Fig.~\ref{fig:inequality}c) shows limited separation among age and sex groups within each city. The radar profiles for each city are nearly regular, indicating similar Deficit Palma Ratio values across children, youth, working-age adults, elderly groups, males, and females. City differences are more visible than demographic differences, with Kuala Lumpur higher and Singapore lower across all groups.

The land-use comparison (Fig.~\ref{fig:inequality}d) shows clearer separation than the demographic comparison. In all four cities, Nature zones have higher Deficit Palma Ratio values than Residential and Commercial zones. Commercial zones are generally lower, while Residential zones occupy an intermediate position. The city ranking remains broadly consistent across land-use groups. In the present analysis, exposure inequality is more clearly separated by city and land-use group than by age and sex group.

\FloatBarrier

\section{Discussion}
\label{sec:discussion}

The main contribution of this study is that it substantially reduces the dependence of city-scale subjective streetscape assessment on SVI by enabling perception estimation from AlphaEarth and urban contextual data. This allows subjective streetscape perceptions to be mapped continuously across entire urban road networks rather than being restricted to locations with available SVI, providing spatially consistent information for city-scale analysis. By integrating perception maps with population data, the framework further identifies disparities in population exposure to perceived environmental quality, enabling comparisons across population-density, demographic and land-use groups. Together, these capabilities provide a practical foundation for evidence-based urban renewal, public-space evaluation, and human-centred urban planning. More broadly, these results suggest that cross-view urban sensing, linking remote sensing observations with subjective human experience of the built environment, holds considerable promise as a scalable paradigm for future urban perception research.

Two limitations should be considered when interpreting the results. First, the AlphaEarth features represent annual remote sensing observations and are therefore suitable for annual perception mapping rather than seasonal or short-term monitoring. This limitation is particularly relevant in mid- and high-latitude regions, where streetscape appearance varies substantially across seasons and would require higher-temporal-resolution remote sensing data \citep{zhao_quantifying_2025}. Second, the perception labels are derived from the pretrained SVI-Percept model, which inherits the characteristics of its original crowdsourced ratings. Consequently, the predicted perception maps should be interpreted within that scoring framework. For applications in regions with different perceptual or cultural contexts, local calibration or fine-tuning could further improve the reliability and transferability of the model \citep{quintana_global_2025}.

The proposed framework is not limited to the five perceptual dimensions examined in this study. Their selection reflects the availability of a consistent perception label system rather than a methodological constraint. As human perception labels for additional dimensions become available, the framework could be extended to predict and map these dimensions. Future work could also explore the integration of large language models (LLMs) and vision-language models (VLMs) to derive semantically meaningful features from remote sensing imagery and auxiliary geospatial data \citep{hou_urban_2025,zhang_earthgpt_2024}. Such features may improve model interpretability and transferability, although their value for perception prediction remains to be systematically validated.

\section{Conclusion}
\label{sec:conclusion}

This study develops a Cross-View Perception Prediction Analysis Framework. Its core model, CVLNet, combines AlphaEarth and urban contextual features to estimate subjective streetscape perception, outperforming baseline models across five perceptual dimensions in four Southeast Asian cities. The trained model generates citywide road-level streetscape perception maps and, combined with WorldPop data, quantifies population exposure inequality across population-density, demographic, and land-use groups. Overall, the proposed framework bridges remote sensing observations and subjective streetscape perception, providing a practical tool for large-scale urban planning, urban renewal, and public-space evaluation.


\section*{CRediT authorship contribution statement}

\textbf{Peilin Li}: Conceptualization, Methodology, Software, Formal analysis, Investigation, Data curation, Writing -- original draft, Visualization.
\textbf{Pengfei Chen}: Conceptualization, Supervision, Writing -- review \& editing, Funding acquisition.
\textbf{Jingyu Wang}: Data curation, Investigation, Software.
\textbf{Zhifeng Yang}: Data curation, Formal Analysis, Investigation.
\textbf{Tiansheng Chen}: Investigation, Visualization.
\textbf{Mengjie Gong}: Data curation, Visualization, Software.
\textbf{Xiao Cheng}: Supervision, Writing -- review \& editing.

\section*{Declaration of competing interest}

The authors declare that they have no known competing financial interests or personal relationships that could have appeared to influence the work reported in this paper.

\section*{Data availability}

The AlphaEarth Foundations data are publicly available from Google DeepMind. WorldPop population data are available at \url{https://www.worldpop.org/}. OpenStreetMap data are available at \url{https://www.openstreetmap.org/}. Google Street View imagery was accessed through the Google Street View API. Code and trained model weights will be made available upon acceptance.

\section*{Acknowledgements}

This research was funded by Guangdong Basic and Applied Basic Research Foundation, grant number 2026A1515012474 and the National Natural Science Foundation of China, grant number 42101351.  
The authors acknowledge the financial support from the Medicine-Engineering Convergence Seed Fund (2025) at Sun Yat-sen University.

\section*{Declaration of generative AI and AI-assisted technologies in the manuscript preparation process}

During the preparation of this work, the authors used ChatGPT in order to polish language and improve readability. After using this tool, the authors reviewed and edited the content as needed and take full responsibility for the content of the published article.


\bibliographystyle{elsarticle-harv}
\bibliography{reference}

\clearpage
\appendix
\setcounter{figure}{0}
\renewcommand{\thefigure}{A.\arabic{figure}}
\section{Supplementary Materials}

\subsection{AlphaEarth Patch-size Selection}
\label{app:patchsize}

\begin{figure}[htbp]
  \centering
  \includegraphics[width=0.9\textwidth]{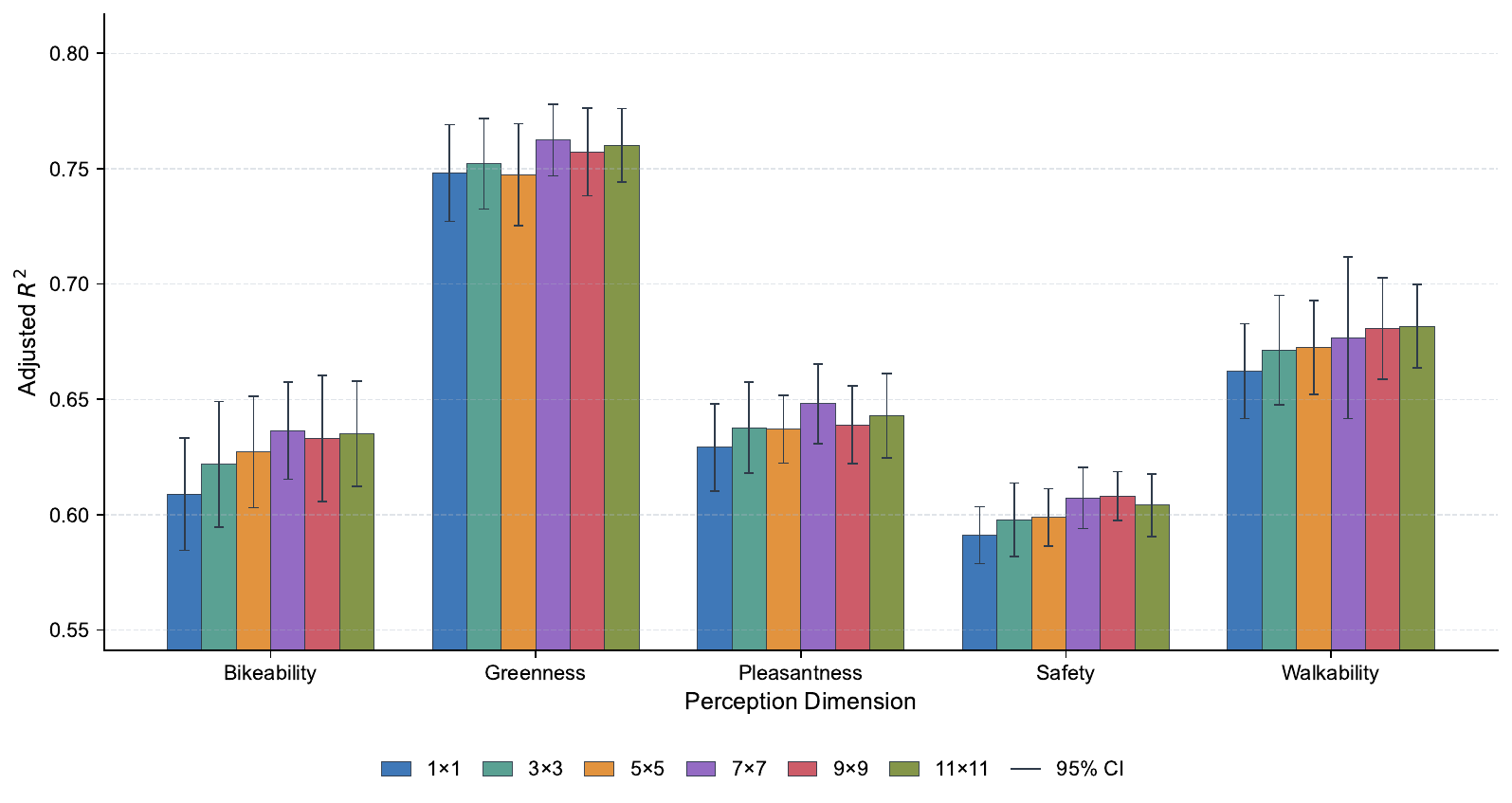}
  \caption{Cross-validation comparison of different AlphaEarth patch sizes.}
  \label{fig:patchsize}
\end{figure}

Different extraction patch sizes were tested for AlphaEarth features (Fig.~\ref{fig:patchsize}). Considering both prediction accuracy and input size, the $7\times7$ patch was selected as the AlphaEarth patch size for CVLNet.

\subsection{SVI Coverage Analysis}
\label{app:svi_coverage}

\begin{figure}[htbp]
  \centering
  \includegraphics[width=0.9\textwidth]{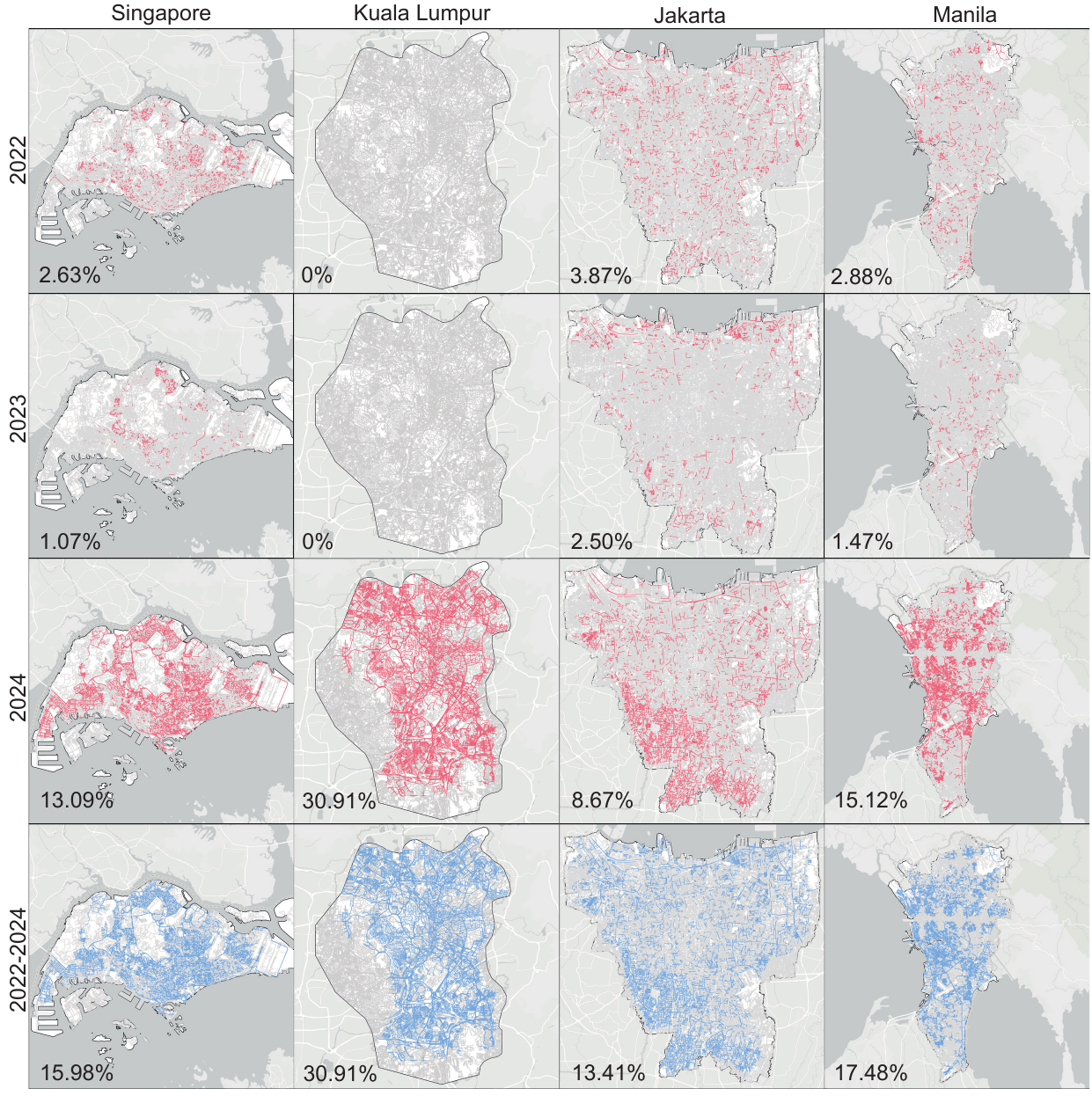}
  \caption{Spatial distribution of street view sampling points across the four study cities.}
  \label{fig:coverage}
\end{figure}

To assess SVI coverage, the availability of Google Street View imagery was checked at points spaced 100\,m apart along the road network. A road segment was considered covered if imagery was available for at least one sampling point on that segment. The coverage ratio was calculated as the total length of covered road segments divided by the total road-network length in each city. As shown in Fig.~\ref{fig:coverage}, SVI coverage varies considerably across the four study cities and is generally limited, with most observations concentrated in urban core areas. Even after aggregating all available street view imagery acquired between 2022 and 2024, the cumulative coverage ratios reach only 30.91\% for Kuala Lumpur, 17.48\% for Manila, 15.98\% for Singapore, and 13.41\% for Jakarta. As a result, street-view-based methods alone cannot generate spatially continuous citywide perception maps.

\end{document}